\documentclass[11pt]{article}

\usepackage[final]{acl}

\usepackage{times}
\usepackage{latexsym}
\usepackage[T1]{fontenc}
\usepackage{ragged2e}
\usepackage{xcolor}
\usepackage{subcaption}
\usepackage{amssymb,amsmath}
\usepackage{booktabs}
\usepackage{array}
\usepackage{enumitem}
\usepackage{listings}
\usepackage[utf8]{inputenc}
\usepackage{microtype}
\usepackage{graphicx}
\graphicspath{{../plots/}{plots/}{}}

\title{Reliable but Design-Sensitive: \\
Instrument Uncertainty in LLM Annotation}

\author{
  Thomas Reiter$^{1}$,
  Christoph Kern$^{2}$,
  Fedor Miasnikov$^{2}$,
  Sofiia Nikolenko$^{2}$, \\
  {\bf Rob Chew}$^{4}$,
  {\bf Stephanie Eckman}$^{1,3}$,
  {\bf Frauke Kreuter}$^{2,3}$ \\
  $^{1}$Amazon \\
  $^{2}$ Ludwig Maximilian University of Munich, Germany\\
  $^{3}$University of Maryland, Social Data Science Center, USA \\
  $^{4}$RTI International, North Carolina, USA \\
  \texttt{txreiter@amazon.com}}

\begin{document}
\maketitle

\begin{abstract}
Large language models (LLMs) can give reliable labels under one setup yet change those labels when researchers make other reasonable design choices. We tested seven LLMs, 12 task designs, three independent runs, and 3,000 tweets labeled for offensive language and hate speech. Repeating the same model and task design produced high agreement (median Fleiss' $\kappa = 0.91$). Agreement fell when we changed the task design for the same tweets (median Cohen's $\kappa = 0.76$). Task design and model choice increased the variance of estimated prevalence by factors of 76.7 for offensive language and 110.6 for hate speech compared with sampling variance alone. Variation across LLM task designs reached 560--572 basis points, compared with 270--331 basis points across five human instrument versions. Confidence scores did not solve this problem. They tracked repeated model outputs more closely than agreement with human labels, and grouping six tweets in one prompt lowered mean offensive-language confidence by 660 basis points. We call the variation caused by task design and model choice \textit{instrument uncertainty}. Researchers can measure it only by comparing reasonable task designs. Repeating one setup or relying on confidence scores cannot replace that test.\footnote{This work was conducted independently of the authors' employment at Amazon and does not reflect the views, opinions, or positions of Amazon.}
\end{abstract}

\section{Introduction}

An LLM can reproduce its labels across repeated runs while producing a different prevalence estimate under another defensible task design. Researchers increasingly use LLMs to generate labels across computational social science and NLP pipelines \citep{gilardi2023chatgpt}, but standard evaluations often compare only one model-design combination with a human benchmark \citep{smolinski2024scaling}. That comparison measures benchmark performance under the selected design, but it does not test whether the labels or resulting estimate survive reasonable changes in task structure, presentation format, confidence elicitation, or model choice.

A recent task force report from the American Association for Public Opinion Research \citep{aapor2026responsible} defines four criteria for evaluating generative AI from a user's perspective: \textbf{validity} (targeting the intended construct), \textbf{performance} (matching a benchmark), \textbf{reliability} (consistency under identical conditions), and \textbf{sensitivity} (stability across defensible task designs). Benchmark performance has a large literature, but sensitivity and reliability remain less examined in automated annotation.

To evaluate sensitivity, the report recommends testing annotation systems under plausible alternative specifications, such as varying prompt wording, decision rules, or model architectures. We execute this test at scale: seven LLMs annotated 3,000 tweets for offensive language (OL) and hate speech (HS) across a fully crossed factorial design over three independent runs.

Our primary finding is that reliability within a task design does not imply stability across task designs. 
This distinction creates a diagnostic trap. A standard evaluation repeats one fixed task design and measures agreement with a human benchmark. That evaluation can establish run-to-run reliability and benchmark performance, but it cannot establish whether the labels survive alternative, defensible task-design choices. Table~\ref{tab:framework} separates these criteria. Our experiment estimates reliability and sensitivity, uses human labels as a performance reference; it does not address the validity of the underlying offensive-language and hate-speech constructs.

\begin{table}[t]
\centering
\footnotesize
\caption{The four evaluation criteria adapted from \citet{aapor2026responsible} and the comparison required for each.}
\label{tab:framework}
\begin{tabular}{ll}
\toprule
\textbf{Criterion} & \textbf{Comparison} \\
\midrule
Validity & Outputs vs. intended construct \\
\addlinespace[2pt]
Performance & Model vs. human benchmark \\
\addlinespace[2pt]
Reliability & Same model-design, repeated runs \\
\addlinespace[2pt]
Sensitivity & Across task designs and models \\
\bottomrule
\end{tabular}
\vspace{-0.35cm}
\end{table}

We make four contributions:
\begin{enumerate}[leftmargin=*,itemsep=2pt,topsep=2pt]
    \item \textbf{Reliability versus sensitivity.} We compare three repeated runs within each model-design cell with labels produced under alternative task designs. Median agreement falls from Fleiss' $\kappa = 0.91$ within designs to Cohen's $\kappa = 0.76$ across designs.
    \item \textbf{Instrument variance.} We identify which task-design and model choices shift labels and decompose their contribution to estimated prevalence. Task design and model choice increase variance by factors of 76.7 for offensive language and 110.6 for hate speech relative to nominal sampling variance.
    \item \textbf{Human benchmarking.} We re-analyze 44,900 human ratings of the same 3,000 tweets across five instrument versions. LLM task-design variation reaches 560--572 basis points, compared with 270--331 basis points for humans.
    \item \textbf{Confidence audit.} We test whether stated confidence identifies labels that change across runs, task designs, and human referents. Confidence tracks repeated model outputs more closely than agreement with humans and changes when identical tweets are presented in batches.
\end{enumerate}

An interactive explorer of all 755,999 LLM labels and 44,900 human ratings is available at are at \url{https://github.com/soda-lmu/explore_labels}.

\section{Related Work}

\paragraph{LLMs as Survey Elicitation Instruments.}
Prompting an LLM closely resembles administering a survey, where changes in question wording, order, and formatting can alter responses \citep{schuman1981questions}. These effects also occur in human annotation and can alter downstream model performance \citep{kern2023annotations, eckman2024position}. \citet{chew2026groundtruth} formalize human annotation as a measurement process and decompose label variation into sources tied to items, annotators, labeling situations, and relationships among annotators. We extend this measurement perspective to LLM annotation by estimating variation introduced by model and task-design choices. Design decisions are therefore part of the measurement instrument rather than implementation details.

\paragraph{Prompt Optimization vs. Task-Design Sensitivity.}
Recent NLP studies examine prompt optimization \citep{atreja2024prompt} and position bias \citep{dominguez2024questioning}, but they generally seek one preferred prompt rather than estimate variation across defensible choices. Batch prompting reduces API costs \citep{cheng2023batch,lin2023batchprompt} but introduces inter-item interference and context sensitivity \citep{larionov2025batchgemba}. \citet{liu2026what} introduced Inter-Prompt Reliability to measure stability across semantically equivalent prompt variations and contrasted interpretive tasks with knowledge-anchored tasks. \citet{kamal2026prompt} found that robustness differs across objective, belief-style, and value-oriented evaluations. \citet{merlo2025costeffective} use random samples of human labels to estimate the agreement and error of LLM-generated relevance judgments with 95\% confidence, requiring human review of 6--16\% of the generated labels. Their framework reduces the cost of benchmark validation but holds the model and prompt fixed, so it does not measure whether results change across reasonable task designs.

\paragraph{Analytic Flexibility in AI Measurement.}
\citet{cummins2025threat} varies model choice, sampling parameters, prompt format, and contextual information across 252 configurations of LLM-generated survey responses. The configurations differ in their recovery of participant rankings, response distributions, and correlations. In a published use case, correlations between human and synthetic association structures range from $r=.23$ to $r=.84$ across 66 configurations. \citet{dell2026measurement} describe the broader measurement problem: AI shifts the constraint from finding one scalable measure to choosing among plausible measures that can support different empirical conclusions. They recommend explicit measurement targets and validation designed around those targets. Both papers identify analytic flexibility as a threat to inference when researchers can choose among AI-generated measures.

\paragraph{Positioning and Contributions.}
Prior studies measure sensitivity to wording, framing, and formatting changes \citep{liu2026what,kamal2026prompt}, while recent work documents analytic flexibility across LLM-generated measures \citep{cummins2025threat,dell2026measurement}. We extend this work with a crossed annotation experiment over task structure, item batching, confidence elicitation, and model selection. The experiment links task-design sensitivity to prevalence uncertainty, compares LLM and human instrument effects, and tests whether confidence identifies labels that change across designs. We call the resulting variation \textit{instrument uncertainty}.

\section{Methods}
\label{sec:methods}

We evaluate reliability and task-design sensitivity using a multi-model annotation experiment on offensive language (OL) and hate speech (HS). Both constructs lack a single gold standard, combine objective and subjective criteria, and are used in human and automated content-moderation benchmarks \citep{sachdeva2022measuring}.

\subsection{Dataset and Models}
We annotated 3,000 tweets from \citet{kern2023annotations}, drawn originally from \citet{davidson2017automated}. We evaluated seven LLMs across three model families and a wide cost range: \textbf{GPT-4o-mini}, \textbf{GPT-5.4}, \textbf{Llama 3.1 8B}, \textbf{Llama 3.1 70B}, \textbf{Llama 4}, \textbf{Mistral Large 3}, and \textbf{Mistral Medium 3.5}. See Appendix~\ref{app:costs} for cost data. We treat these models as a sample from the space of deployed annotators to estimate model-level variance. All labels and analysis scripts are available at
\url{https://github.com/T-Reiter/llm-annotation-sensitivity}.

\subsection{Factorial Experimental Design}
We varied the annotation instrument using a $3 \times 2 \times 2$ factorial design, yielding 12 task designs:
\begin{enumerate}[leftmargin=*,itemsep=1pt,topsep=2pt]
    \item \textbf{Task Structure (3 levels):} OL and HS were elicited jointly in one API call (\textit{OL first} or \textit{HS first}) or separately in independent calls.
    \item \textbf{Presentation Format (2 levels):} Tweets were presented individually or in fixed, randomly assigned batches of six.
    \item \textbf{Confidence Elicitation (2 levels):} The task requested labels alone or labels with a confidence score ($0\text{--}100\%$).
\end{enumerate}

Every condition represents a defensible task-design choice. Base definitions and instructions were held constant. See Appendix~\ref{app:sample_prompts} for the prompts. 

Each tweet received three independent runs per condition at temperature $1$, producing 755,999 labels each for HS and OL.\footnote{One of 756,000 responses was unusable: Llama 3.1 8B, joint labeling with OL asked first, individual presentation, no confidence, run 1. We drop this observation.}

\subsection{Analytic Strategy}
\label{sec:analytic_strategy}

We ask whether labels change across runs and task designs, which choices drive those changes, how much variance they add to prevalence, how that variance compares with human instrument effects, and whether confidence identifies unstable labels. The analysis moves from detecting sensitivity to explaining its sources, measuring its consequences, benchmarking its magnitude, and testing confidence as a diagnostic.

\paragraph{Reliability Within Designs and Sensitivity Across Designs.}
We first separate two forms of stability that are often treated as the same. Within each model-design cell, we measure run-to-run reliability using Fleiss' $\kappa$ across the three independent runs \citep{fleiss1971measuring}. This comparison asks whether a fixed model-design combination reproduces its labels when only the run changes.

High run-to-run reliability does not establish that labels are stable across task designs. We therefore take the modal label across the three runs and calculate Cohen's $\kappa$ between pairs of task designs within the same model. Cohen's $\kappa$ then asks whether labels persist when the researcher changes the task structure, confidence request, or presentation format. Together, the two statistics distinguish reproducibility under a fixed instrument from sensitivity to the choice of instrument.

\paragraph{Sources and Magnitude of Instrument Variance.}
The $\kappa$ comparisons detect sensitivity but do not identify its source or consequence. We therefore fit three models in sequence. A pooled item-level model estimates average task-design effects, and a varying-slope model tests whether those effects differ across LLMs. A crossed prevalence model then decomposes variation from runs, task design, model choice, and Model $\times$ Design interactions. Comparing these components with nominal sampling variance measures how much uncertainty a sampling-only calculation misses.

We begin with a pooled linear probability model, estimated separately for OL and HS.\footnote{Appendix~\ref{app:main_table} reports mixed-effects logistic models as a specification check.} Let $y_i \in \{0,1\}$ denote the label for observation $i$, $j[i]$ denote its tweet, and $m[i]$ denote its LLM. The pooled specification is:
\begin{equation*}
\begin{aligned}
y_i &= \alpha_{j[i]} + \mathbf{x}_i^\top\boldsymbol{\beta}
      + \mathbf{w}_i^\top\boldsymbol{\gamma} + \epsilon_i, \\
\alpha_j &\sim N(\mu_\alpha,\sigma_\alpha^2),
\qquad
\epsilon_i \sim N(0,\sigma_y^2),
\end{aligned}
\label{eq:pooled_lpm}
\end{equation*}
where $\mathbf{x}_i$ contains four indicators for task structure, confidence elicitation, and batching, and $\mathbf{w}_i$ contains model indicators. The varying intercept $\alpha_{j[i]}$ accounts for repeated labels of the same tweet. The coefficients $\boldsymbol{\beta}$ estimate average task-design effects, and $\boldsymbol{\gamma}$ estimates model differences after holding task design constant.

The pooled model assumes that each task-design effect is constant across models. We therefore fit a varying-slope model that allows each effect to differ by model:
\begin{eqnarray*}
y_i = \alpha_{j[i]} + \mathbf{w}_i^\top\boldsymbol{\gamma}
      + \sum_{k=1}^{4} x_{ik}\beta_{k,m[i]} + \epsilon_i, \\
\beta_{km} = \beta_k + b_{km},
\qquad
b_{km} \sim N(0,\sigma_{\beta_k}^2), \\
\operatorname{Cov}(b_{km},b_{\ell m}) = 0
\qquad \text{for } k \neq \ell .
\label{eq:varying_slope_lpm}
\end{eqnarray*}
Here, $\beta_k$ is the average effect of factor $k$, and $b_{km}$ is model $m$'s deviation from that average. Separate variance components for the four uncorrelated slopes measure cross-model differences in each task-design effect.

To measure consequences for prevalence, we calculate one estimate for each model, task design, and run, yielding 252 estimates per outcome ($7$ models $\times$ $12$ designs $\times$ $3$ runs). Let $\widehat{p}_{mdr}$ denote the prevalence estimate for model $m$, task design $d$, and run $r$. We fit the crossed random-effects model:
\begin{equation*}
\begin{aligned}
\widehat{p}_{mdr} &= \mu + \eta_m + \delta_d + \zeta_{md} + \epsilon_{mdr}, \\
\eta_m &\sim N(0,\sigma_{\mathrm{model}}^2),\\
\delta_d &\sim N(0,\sigma_{\mathrm{design}}^2), \\
\zeta_{md} &\sim N(0,\sigma_{\mathrm{model}\times\mathrm{design}}^2),\\
\epsilon_{mdr} &\sim N(0,\sigma_{\mathrm{run}}^2).
\end{aligned}
\end{equation*}
The last four terms capture variation from model, task design, their interaction, and repeated runs. The interaction captures the same task design affecting models differently. 

Choosing a task design exposes an application to both its average effect, $\delta_d$, and its model-specific effect, $\zeta_{md}$. We therefore define task-design variance as:
\begin{equation*}
\sigma_{\mathrm{task\ design}}^2
=
\sigma_{\mathrm{design}}^2
+
\sigma_{\mathrm{model}\times\mathrm{design}}^2.
\end{equation*}
Task-design variance combines the average design and Model $\times$ Design components because choosing a design exposes an application to both.

To compare instrument variance with sampling variance, we calculate nominal binomial sampling variance as $\bar{p}(1-\bar{p})/3{,}000$ and add the run, task-design, and model components sequentially. We define the design effect \citep[$deff$,][]{kish1965survey} as cumulative variance divided by nominal sampling variance. A $deff$ of 1 indicates no added variance; values above 1 report how many times larger cumulative variance is than sampling variance alone.

\paragraph{Human Benchmarking.}
The variance decomposition measures LLM instrument uncertainty, but it does not tell us whether its magnitude is specific to LLMs. We therefore re-analyze 44,900 human ratings collected by \citet{kern2023annotations} on the same 3,000 tweets across five instrument versions. Holding the items and constructs constant provides a human reference for how much prevalence changes when the instrument changes.

We compare the standard deviation of prevalence across human instrument versions with the task-design standard deviation across LLM task designs. We also compare the corresponding human and LLM design effects. These comparisons do not treat the human labels as ground truth. Instead, they place human and LLM task-design sensitivity on the same prevalence scale and ask whether the LLM estimates are more sensitive to instrument choice than the human estimates.


\paragraph{Confidence Audit.}
The preceding analyses measure instrument sensitivity after labels have been collected. We next ask whether the model's stated confidence can identify labels that are likely to change under another task design. If confidence captures this form of uncertainty, labels assigned lower confidence should have lower agreement across task designs, and stated confidence should match the observed rate of cross-design agreement.

The six confidence-eliciting conditions provide stated confidence scores from 0\% to 100\%. Our primary agreement target is the same model labeling the same tweet under another task design. This target tests whether confidence tracks task-design sensitivity. We use three additional targets to determine what confidence tracks instead. Agreement with another run of the same design measures confidence in run-to-run reproducibility. Agreement with an individual human annotator measures correspondence with one human judgment. Agreement with the human majority label measures correspondence with the aggregate human benchmark. Comparing the four targets separates confidence in self-reproduction, confidence in stability across task designs, and confidence in human agreement.

For each agreement target $r$, let $c_i \in [0,1]$ denote the model's stated confidence for item $i$, and let $a_{ir}=1$ if the model's label agrees with target $r$ and $a_{ir}=0$ otherwise. We divide the confidence scale into ten equal-width bins, $\mathcal{B}_1,\ldots,\mathcal{B}_{10}$. Binning allows us to compare stated confidence with the observed agreement rate among labels assigned similar confidence. We calculate Expected Calibration Error (ECE) as \citep{guo2017calibration}:
\begin{equation}
\operatorname{ECE}_r
=
\sum_{b=1}^{10}
\frac{n_b}{N}
\left|
\bar{c}_b-\bar{a}_{br}
\right|,
\label{eq:ece}
\end{equation}
where $n_b=|\mathcal{B}_b|$, $N=\sum_b n_b$, $\bar{c}_b=n_b^{-1}\sum_{i\in\mathcal{B}_b}c_i$ is mean stated confidence in bin $b$, and $\bar{a}_{br}=n_b^{-1}\sum_{i\in\mathcal{B}_b}a_{ir}$ is observed agreement with target $r$ in that bin. The weights $n_b/N$ give each bin influence in proportion to its share of the evaluated labels, rather than giving a sparse bin the same influence as a common bin.

An ECE near 0 means that stated confidence matches observed agreement with the specified target. It does not establish that the target is correct or that confidence has the same meaning across targets. We therefore name the agreement target for every ECE estimate. Comparing ECE across the four targets tells us whether stated confidence is calibrated most closely to repeated execution, another task design, an individual human, or the human majority.

\section{Results}
\label{sec:results}

Appendix~\ref{app:additional_results} reports the full regression, agreement, confidence, and cost results.

\subsection{Reliable Within Designs, Sensitive Across Designs}
\label{sec:reliability_sensitivity}

\begin{figure}[t]
    \centering
    \includegraphics[width=\linewidth]{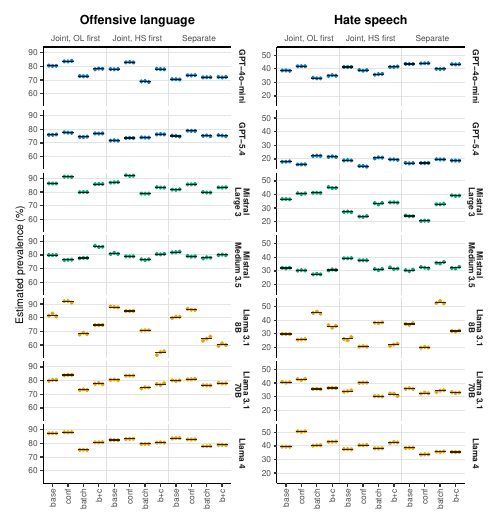}
    \caption{\textbf{Prevalence changes more across task designs and models than across repeated runs.} Each cluster contains three runs for one model and task design. The same 3,000 tweets were used in every combination.}
    \label{fig:prev_stability}
\end{figure}

Figure~\ref{fig:prev_stability} shows the offensive-language and hate-speech prevalence estimates for each model, task design, and run. The three runs produce similar estimates when the model and task design stay fixed. The estimates differ much more when the task design or model changes. For hate speech, prevalence ranges from 14.5\% to 53.0\% across the 84 model-design combinations. 

Kappa measures label agreement after accounting for agreement expected by chance. Fleiss' $\kappa$ compares the three runs within the same model and task design. Across the 166 cells with defined values, the median Fleiss' $\kappa$ is 0.91 (Table~\ref{tab:agreement_ladder}). Six of the seven models have mean within-design agreement above $\kappa = 0.79$. Llama 3.1 8B has lower within-design agreement than the other models: Fleiss' $\kappa$ ranges from 0.19 to 0.80. 

Cohen's $\kappa$ compares labels after the task design changes while the model and tweets stay fixed. Median agreement across task designs is $\kappa = 0.76$. Median agreement with the human majority label is lower, at $\kappa = 0.54$. Every model follows the same order in Table~\ref{tab:agreement_ladder}: agreement is highest across repeated runs, lower across task designs, and lowest against the human majority label.

A fixed model and task design can therefore produce consistent labels across repeated runs while remaining sensitive to choices that change both the labels and estimated prevalence.

\subsection{Task Design and Model Choice Create Instrument Variance}
\label{sec:instrument_variance}
The mixed-effects models identify which task-design and model choices generate the across-design spread. Tweet content accounts for 59.1\% of OL label variance and 56.3\% of HS label variance in the pooled linear probability models (Table~\ref{tab:taskeffects_all}). 

Model choice produces the largest mean shifts. Across models, HS prevalence ranges from 18.6\% for GPT-5.4 to 39.8\% for GPT-4o-mini, a difference of 2,120 basis points on identical tweets. After controlling for task design, model coefficients span 2,120 basis points relative to GPT-4o-mini. Adding model identity raises marginal $R^2$ from 0.007 to 0.013 for OL and from 0.001 to 0.021 for HS.

Task-design choices also move the decision threshold. Batch presentation lowers OL prevalence by 590 basis points. Joint labeling with OL asked first raises OL prevalence by 270 basis points and HS prevalence by 230 basis points. Requesting confidence raises OL prevalence by 230 basis points and lowers HS prevalence by 100 basis points. Asking for confidence therefore changes the labels whose uncertainty it is intended to measure.

\begin{figure}[t]
    \centering
    \includegraphics[width=\columnwidth]{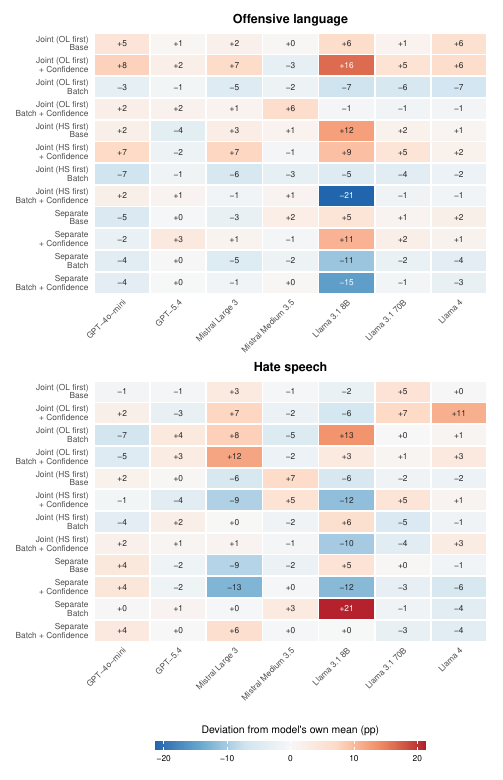}
    \caption{\textbf{Task-design effects differ in direction and magnitude across models.} Each cell reports the deviation from that model's mean prevalence for one task design. Shared row patterns indicate average task-design effects; differences within a row reveal Model $\times$ Design interactions. Batching and confidence elicitation do not shift all models in the same direction.}
    \label{fig:desc_all}
\end{figure}

These average effects do not transport across models. Adding per-model random slopes improves fit for both OL ($\chi^2_4 = 13{,}522$, $p < .001$) and HS ($\chi^2_4 = 17{,}817$, $p < .001$). One pooled effect therefore conceals model-specific changes in direction and magnitude. Batching raises HS prevalence in three models and lowers it in four (Table~\ref{tab:permodel}). A task-design adjustment that reduces prevalence for one model can increase it for another. Figure~\ref{fig:desc_all} shows each task design's deviation from the model-specific mean prevalence for OL and HS.

\begin{table*}[t]
\centering
\footnotesize
\setlength{\tabcolsep}{4pt}
\caption{\textbf{Cumulative uncertainty in estimated label prevalence.} Component standard deviations are estimated from crossed random-effects models of run-level prevalence. Task-design variation combines the task-design and Model $\times$ Design variance components. Cumulative standard deviations add independently estimated components as variances. The 95\% column reports $1.96 \times \mathrm{SD}_{\mathrm{cum}}$, and $deff$ is cumulative variance divided by nominal binomial sampling variance. Standard deviations and interval half-widths are reported in basis points.}
\label{tab:deff}
\begin{tabular}{@{}lrrrrrrrr@{}}
\toprule
& \multicolumn{4}{c}{\textbf{Offensive language}}
& \multicolumn{4}{c}{\textbf{Hate speech}} \\
\cmidrule(lr){2-5}\cmidrule(lr){6-9}
Variance source added
& Component SD & SD$_{\mathrm{cum}}$ & $\pm$95\% & $deff$
& Component SD & SD$_{\mathrm{cum}}$ & $\pm$95\% & $deff$ \\
\midrule
Sampling only (nominal)
& 75 & 75 & $\pm$150 & 1.00
& 86 & 86 & $\pm$170 & 1.00 \\
Run-to-run variation
& 36 & 83 & $\pm$160 & 1.23
& 38 & 94 & $\pm$180 & 1.19 \\
Task design, incl. Model $\times$ Design
& 560 & 567 & $\pm$1,110 & 57.7
& 572 & 579 & $\pm$1,140 & 45.5 \\
Model choice
& 325 & 653 & $\pm$1,280 & 76.7
& 693 & 903 & $\pm$1,770 & 110.6 \\
\bottomrule
\end{tabular}
\smallskip
\begin{minipage}{\textwidth}
\raggedright
\footnotesize \textit{Note}: Estimates use 755,999 labels from seven LLMs and twelve task designs over three runs on 3,000 tweets. Nominal sampling error assumes a binomial distribution at $n=3{,}000$. The survey design effect ($deff$) is total empirical variance, including sampling, run-to-run, task design, and model choice, divided by nominal binomial sampling variance.
\end{minipage}
\end{table*}

Table~\ref{tab:deff} translates these shifts into cumulative uncertainty. For OL, nominal sampling alone produces a 95\% interval half-width of 150 basis points. Adding run-to-run variation changes the half-width to 160 basis points. Adding task-design and Model $\times$ Design variation raises it to 1,110 basis points. Adding model choice raises it to 1,280 basis points and yields $deff = 76.7$. For HS, the 95\% interval half-width, accounting for all sources of variability, is 1,770 basis points, with $deff = 110.6$. 

\subsection{LLM Task-Design Variation Exceeds Human Task-Design Variation}
\label{sec:humans}
Human labels of the same tweets also show variability across five instrument versions: the standard deviation on prevalence is 331 basis points for OL and 270 basis points for HS \citep{kern2023annotations}. None of the versions used by human labelers asked for confidence scores, so we judge these results against the six LLM conditions that did not ask for confidence scores. Across these six LLM task designs, the corresponding standard deviations are 463 basis points for OL and 513 basis points for HS. LLM task-design variation is therefore 1.4 times the human value for OL and 1.9 times the human value for HS, although the versions were not the same for the two types of labelers.

Similarly, human instrument versions produce design effects of 13--15, compared with 76.7 for OL and 110.6 for HS when LLM task design and model choice are included. 

Presenting OL and HS jointly lowers human OL labeling by 598 basis points but raises LLM OL labeling by 267 basis points (Table~\ref{tab:taskeffects_all}). 

\subsection{Stated Confidence Does Not Measure Stability Across Task Designs}
\label{sec:confidence}
In the six confidence-eliciting conditions, the models returned a usable confidence score for 98.1\% of labels. The missing scores are concentrated in the  batched presentation design (97.2\% of OL scores with missing confidence), and in Llama 3.1 8B (73.5\% of all scores with missing confidence). The calibration estimates therefore condition on returning a valid score and give more weight to individual-presentation conditions.

Confidence is calibrated most closely to the model's repeated outputs. Expected Calibration Error ranges from 0.074 to 0.096 against another run of the same design and from 0.050 to 0.059 against another task design. Calibration is worse against an individual human annotator, with ECE values of 0.218 for OL and 0.163 for HS. Against the human majority label, ECE is 0.160 for OL and 0.111 for HS. Panel A of Table \ref{tab:confidence_permodel} reports the pooled ECE estimates. Panel B reports model-specific estimates for the first three agreement targets.

\begin{figure*}[t]
    \centering
    \includegraphics[width=\textwidth]{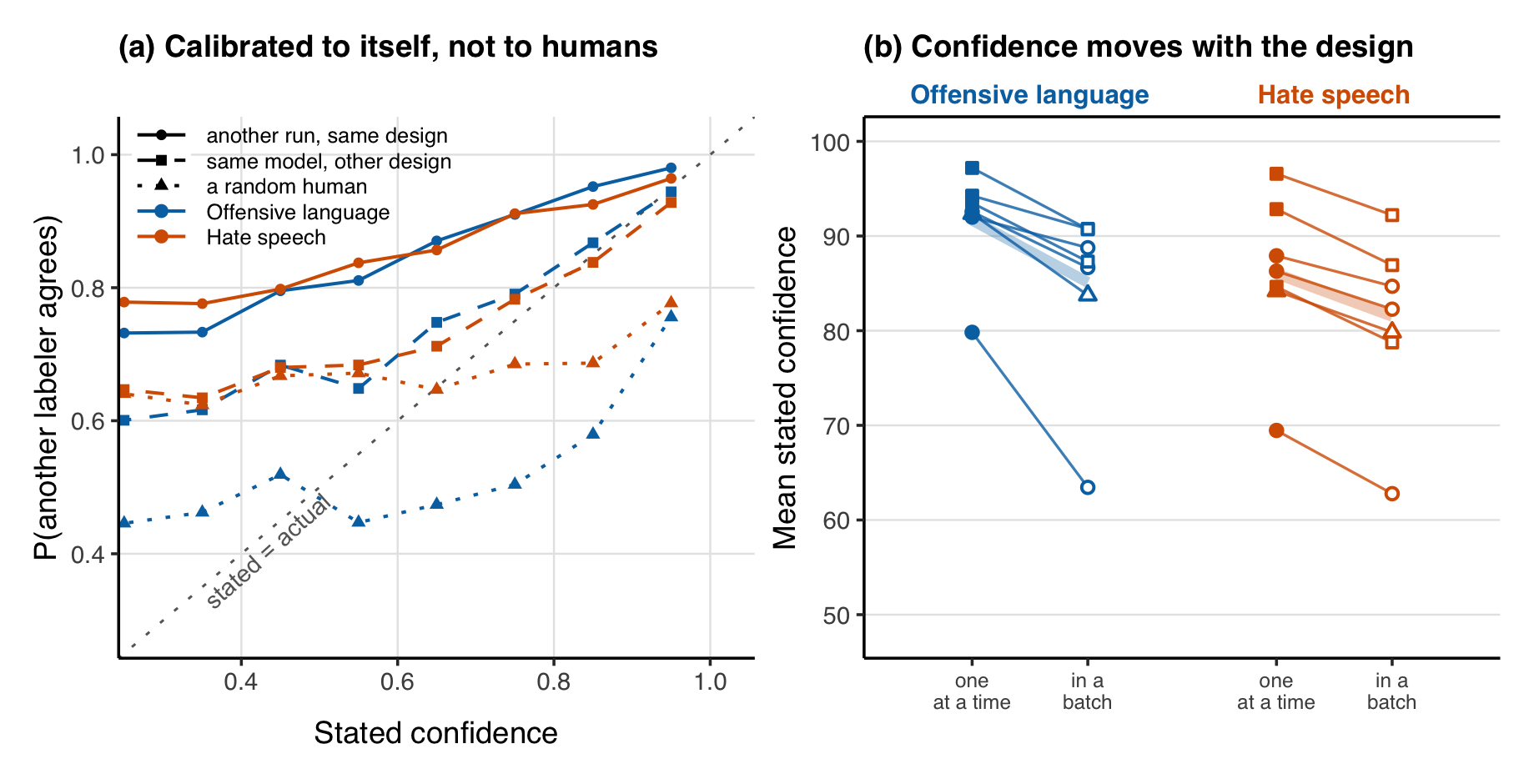}
    \caption{\textbf{Stated confidence tracks repeated model outputs more closely than external agreement and changes with presentation format.} (a) Observed agreement against stated confidence for another run, another task design, and an individual human annotator. The diagonal marks perfect calibration. (b) Mean stated confidence under individual and batched presentation. All seven models report lower confidence when the same tweets are presented six per call, showing that confidence depends on the instrument used to elicit it.}
    \label{fig:confidence}
\end{figure*}

All seven models report lower confidence when the same tweets are presented in batches of six rather than individually (Figure~\ref{fig:confidence}b). 

\section{Discussion}
\label{sec:discussion}

\paragraph{Reliability and Sensitivity Answer Different Questions.}
Repeating one fixed model-design combination establishes whether that combination reproduces itself. It does not establish whether the labels survive another defensible task design. Median agreement falls from Fleiss' $\kappa = 0.91$ within designs to Cohen's $\kappa = 0.76$ across designs. An evaluation based only on repeated runs can therefore describe an annotator as reliable while missing changes that alter the study's prevalence estimate. Reliability should be tested by repeating the same model and design. Sensitivity should be tested by changing the design while holding the items and model fixed.

\paragraph{Instrument Uncertainty Belongs in the Reported Error Budget.}
Task design and model choice produce variance that nominal sampling intervals omit. For 3,000 tweets, the nominal OL interval has a half-width of 150 basis points. Including run, task-design, and model variation increases that half-width to 1,280 basis points. The corresponding HS half-width reaches 1,770 basis points. We call this additional variation \textit{instrument uncertainty} because it arises from the instrument used to elicit the labels.

Instrument uncertainty differs from item ambiguity and parameter uncertainty. Item ambiguity concerns whether a tweet supports a clear label. Parameter uncertainty concerns what a fitted model has learned. Instrument uncertainty concerns which model and task design the researcher selected. Collecting more items under one fixed design reduces sampling error but does not reveal the spread across untested designs. Selecting a larger model does not remove this requirement because the direction and size of the task-design effects differ across the seven models.

\paragraph{Human Task-Design Effects Do Not Predict LLM Effects.}
Human experiments provide a useful reference but cannot substitute for testing the deployed LLM. Among conditions that did not request confidence, LLM task-design variation is 1.4 times the human value for OL and 1.9 times the human value for HS. Joint presentation also moves the two types of annotators in opposite directions: it lowers human OL labeling by 598 basis points but raises LLM OL labeling by 267 basis points. Human pretesting can identify risks in a construct or its wording, but it cannot determine how a specific model-design combination will respond. Researchers must test that combination directly.

\paragraph{Confidence Is Part of the Instrument.}
A confidence score has no target-independent interpretation. Its meaning depends on whether it is expected to represent agreement with another run, another task design, or a human label. In our experiment, stated confidence is calibrated more closely to model agreement than to agreement with an individual human. Confidence can therefore describe consistency relative to a specified target, but it does not provide a general measure of label quality or stability.

Confidence elicitation also changes the instrument it is intended to evaluate. Requesting confidence raises OL prevalence by 230 basis points. Presenting the same tweets in batches of six lowers mean OL confidence by 660 basis points across the seven models. Confidence scores therefore depend on the task design that produces them. Filtering low-confidence labels cannot replace an explicit comparison across task designs.

\paragraph{Implications for Applied Annotation.}
Researchers estimating absolute prevalence should conduct a sensitivity audit before treating an LLM-generated estimate as stable. A minimum audit compares at least two defensible task designs while holding the items and model fixed. The comparison should vary choices that could change the decision boundary, including task structure, batching, and confidence elicitation. Researchers should also repeat the audit across candidate models because the same design change can move different models in opposite directions. The companion explorer runs this audit on our data. It compares any two of the 90 setups and shows which tweets they label differently.

Two task designs can reveal sensitivity, but they cannot estimate the full range of instrument uncertainty. Studies that depend on an absolute prevalence estimate should evaluate a broader set of defensible designs and report the resulting distribution or range. Annotation reports should document the model, model version, prompt, task structure, presentation format, confidence request, temperature, and collection date as measurement parameters rather than implementation details.

Comparative analyses may be less exposed when every group, period, and condition uses the same fixed instrument. This protection requires the instrument effect to remain constant across groups. If task design interacts with group characteristics or time, the resulting differences can still reflect the instrument rather than the underlying construct. Researchers should test this assumption before interpreting group differences, trends, or treatment effects.

\section{Conclusion}\label{sec:conclusion}

LLM annotation systems can be reliable within one task design and sensitive across task designs. Across seven models, twelve designs, three runs, and 3,000 tweets, median agreement falls from Fleiss' $\kappa = 0.91$ within designs to Cohen's $\kappa = 0.76$ across designs. Task design and model choice produce design effects of 76.7 for offensive language and 110.6 for hate speech. Their task-design variation exceeds variation across five human instrument versions, and stated confidence does not recover the missing stability test.

These findings define \textit{instrument uncertainty}: variation introduced by the model and task design used to elicit labels. Researchers cannot estimate this uncertainty by repeating one model-design combination, collecting more items under that combination, or filtering labels by stated confidence. They must compare defensible task designs, report the resulting spread, and include task-design and model settings in the study's measurement description.

\section{Limitations}
\label{sec:limitations}

\paragraph{Task and Construct Scope.}
Our evaluation focuses on offensive language and hate speech detection. These subjective, safety-aligned constructs expose task-design sensitivity, but the estimates may not transfer to objective NLP tasks with tight ground-truth constraints, such as syntactic parsing or extractive question answering.

\paragraph{Task-Design Parameter Space.}
We evaluated twelve task designs across batching, task structure, and confidence elicitation. We did not evaluate few-shot exemplar selection, chain-of-thought reasoning, or multi-turn agent workflows. These choices may introduce additional instrument uncertainty.

\paragraph{Model Selection.}
Our benchmark covers seven open and commercial LLMs available during the study window. The sample excludes model families including Claude and Gemini and is not randomly drawn. The estimated model variance therefore applies to the evaluated set and may understate variation across the full model population.

\paragraph{Prompt Wording.}
We held wording constant and varied structural features. Sensitivity to surface-level wording \citep{sclar2023quantifying} is an additional component that this experiment does not estimate.

\paragraph{Temperature and Time.}
We fixed temperature at 1 and collected labels within one window. Other temperature settings could change reliability and task-design effects. Model behavior can also change behind a stable endpoint. Estimating temporal reliability would require repeated collection on a fixed schedule. The reported design effects exclude this additional source of instrument uncertainty.

\paragraph{No Downstream Model Training.}
We look at how task design changes the labels themselves. We do not train any classifiers on those labels. So we do not know if instrument uncertainty also shows up in a trained model's performance or outputs. Previous work compared task-design effects on human ratings \citet{kern2023annotations} (Section~\ref{sec:humans}). Future work should train models on labels from different task designs and test whether the differences we find here carry over to model performance.

\bibliography{custom}

\newpage
\appendix

\clearpage
\section{Example Prompts and Model Outputs}
\label{app:sample_prompts}
This appendix presents examples of prompts used for LLM annotation in the Base and Batch-Confidence conditions.

\subsection{Base Condition}
In the Base condition, the LLM annotated one tweet for OL and HS jointly, with OL asked first.

\subsubsection*{System Prompt}
\begin{lstlisting}
You are given tweets and your task is to tell whether they contain offensive language (OL) and/or hate speech (HS).

We define hate speech as:
Language that is used to expresses hatred towards a targeted group or language intended to be derogatory, to humiliate, or to insult the members of the group.

[definitions and examples omitted for brevity; see supplementary materials]

First, provide a label for OL, then provide a label for HS.
Use the labels OL and HS if the tweets contain OL or HS.
Use the labels NO and NH if they do not.

Example Input:
Guess who just got an apartment in downtown Columbus? That's right bitch you guessed it, this guy.

Example Output:
OL, NH
\end{lstlisting}

\subsubsection*{User Prompt}
\begin{lstlisting}
Label this tweet:
Fuck that shit I'm out this bitch
\end{lstlisting}

\subsubsection*{Model Response}
\begin{lstlisting}
OL, NH
\end{lstlisting}

\subsection{Batch-Confidence Condition}
In the Batch-Confidence condition, the LLM annotated multiple tweets jointly with confidence scores.

\subsubsection*{System Prompt}
\begin{lstlisting}
You are given multiple tweets and your task is to tell whether each tweet contains offensive language (OL) and/or hate speech (HS).

[identical definitions as above]

For each tweet, provide a label for OL, then provide a label for HS.
For each label also provide a score indicating your confidence in percent.
0% represents lowest possible confidence, 100% represents highest possible confidence.
Separate the labels and confidence scores for the different tweets with ; and start a new line

Example Input:
Guess who just got an apartment in downtown Columbus? That's right bitch you guessed it, this guy.
Subtweet me one more time, you dirty chink whore

Example Output:
OL100, NH80;
OL90, HS90
\end{lstlisting}

\subsubsection*{User Prompt}
\begin{lstlisting}
Label these tweets:

1) 4 old hoes, a ponk, and a dyke outchea having "girl talk"
2) @###### trash
3) RT @###### These hoes be twerkin for Birkin
4) @###### pay your taxes teabaggers
5) @###### Oh yeah. And the annoying damn duck calls?
6) RT @###### Ladies If a nigga always wanna go out and party every weekend. He basically side hoe shopping
\end{lstlisting}

\subsubsection*{Model Response}
\begin{lstlisting}
OL90, NH85;
OL80, NH80;
OL85, NH75;
OL75, NH80;
NO100, NH90;
OL95, HS90
\end{lstlisting}

\section{Additional Results}
\label{app:additional_results}

\subsection{Pooled Regression Tables}
\label{app:main_table}
Table~\ref{tab:taskeffects_all} reports the mixed-effects linear probability models underlying the main-text instrument-variance analysis. Table~\ref{tab:task_structure_glmer} repeats the analysis as mixed-effects logistic regressions.

\begin{table*}[t]
\centering
\caption{Pooled mixed-effects LPMs for OL and HS across three nested specifications. Task estimates task-design effects pooled across models; Models adds model identity; Random Slopes allows task-design effects to vary by model.}
\resizebox{\textwidth}{!}{%
\begin{tabular}{lcccccc}
\toprule
Term & OL Task & OL Models & OL Random Slopes & HS Task & HS Models & HS Random Slopes \\
\midrule
(Intercept) & 0.794*** & 0.765*** & 0.722*** & 0.322*** & 0.389*** & 0.434*** \\
\multicolumn{7}{l}{\textbf{Task Design Effects}} \\
Joint: OL first (vs. Separate) & 0.027*** & 0.027*** & 0.027 & 0.023*** & 0.023*** & 0.023 \\
Joint: HS first (vs. Separate) & 0.011*** & 0.011*** & 0.011 & $-$0.007*** & $-$0.007*** & $-$0.007 \\
With Confidence (vs. Without) & 0.023*** & 0.023*** & 0.023* & $-$0.010*** & $-$0.010*** & $-$0.010 \\
Batch Prompt (vs. Individual) & $-$0.059*** & $-$0.059*** & $-$0.059 & 0.018*** & 0.018*** & 0.018 \\
\multicolumn{7}{l}{\textbf{Model Fixed Effects} (vs. GPT-4o mini)} \\
GPT-5.4 & -- & $-$0.006*** & 0.029*** & -- & $-$0.212*** & $-$0.264*** \\
Mistral Large 3 & -- & 0.087*** & 0.109*** & -- & $-$0.065*** & $-$0.192*** \\
Mistral Medium 3.5 & -- & 0.037*** & 0.069*** & -- & $-$0.071*** & $-$0.094*** \\
Llama 3.1 8B & -- & $-$0.005*** & 0.108*** & -- & $-$0.077*** & $-$0.072*** \\
Llama 3.1 70B & -- & 0.030*** & 0.080*** & -- & $-$0.042*** & $-$0.079*** \\
Llama 4 & -- & 0.058*** & 0.110*** & -- & $-$0.002 & $-$0.085*** \\
\multicolumn{7}{l}{\textbf{Cross-Model Variance Components (Random Slopes)}} \\
Var: Joint OL first & -- & -- & 0.0008 & -- & -- & 0.0037 \\
Var: Joint HS first & -- & -- & 0.0005 & -- & -- & 0.0017 \\
Var: With Confidence & -- & -- & 0.0003 & -- & -- & 0.0027 \\
Var: Batch Prompt & -- & -- & 0.0046 & -- & -- & 0.0037 \\
\multicolumn{7}{l}{\textbf{Random Effects}} \\
$\tau_{00}$\textsubscript{tweet\_id} & 0.098 & 0.098 & 0.098 & 0.125 & 0.125 & 0.125 \\
\midrule
ICC & 0.591 & 0.595 & 0.599 & 0.563 & 0.575 & 0.580 \\
Marginal $R^2$ & 0.007 & 0.013 & 0.016 & 0.001 & 0.021 & 0.031 \\
Conditional $R^2$ & 0.594 & 0.600 & 0.606 & 0.564 & 0.583 & 0.593 \\
AIC & 128866.3 & 116965.3 & 103450.9 & 396112.7 & 361300.8 & 343491.6 \\
BIC & 128947.1 & 117115.2 & 103647.0 & 396193.5 & 361450.8 & 343687.7 \\
LRT $\chi^2_{4}$ (vs. Models) & -- & -- & 13522.37 & -- & -- & 17817.24 \\
LRT $p$ & -- & -- & $<$.001 & -- & -- & $<$.001 \\
\bottomrule
\end{tabular}%
}
\smallskip
\begin{minipage}{\textwidth}
\raggedright
\footnotesize \textit{Note}: $N$\textsubscript{tweet\_id} = 3,000; observations = 755,999. *** $p<.001$, ** $p<.01$, * $p<.05$. Intercept = separate labeling, individual presentation, no confidence, GPT-4o-mini in the Models and Random Slopes columns. Random Slopes adds uncorrelated per-model random slopes for each task-design factor. Models are fitted by REML; AIC, BIC, and the LRT use maximum-likelihood refits.
\end{minipage}
\label{tab:taskeffects_all}
\end{table*}

\begin{table*}[t]
\centering
\caption{Pooled mixed-effects logistic regressions for OL and HS across four nested specifications: Task, Models, Task Interaction, and Design Interaction.}
\resizebox{\textwidth}{!}{%
\begin{tabular}{lcccccccc}
\toprule
Term & OL Task & OL Models & OL Task Int. & OL Design Int. & HS Task & HS Models & HS Task Int. & HS Design Int. \\
\midrule
(Intercept) & 2.608*** & 2.234*** & 2.147*** & 1.998*** & $-$1.786*** & $-$1.116*** & $-$1.169*** & $-$1.004*** \\
\multicolumn{9}{l}{\textbf{Task Design Effects}} \\
Joint: OL first (vs. Separate) & 0.401*** & 0.417*** & 0.711*** & 0.432*** & 0.235*** & 0.257*** & 0.259*** & 0.265*** \\
Joint: HS first (vs. Separate) & 0.156*** & 0.162*** & 0.162*** & 0.167*** & $-$0.075*** & $-$0.083*** & 0.073*** & $-$0.085*** \\
With Confidence (vs. Without) & 0.338*** & 0.351*** & 0.352*** & 0.645*** & $-$0.105*** & $-$0.115*** & $-$0.115*** & 0.220*** \\
Batch Prompt (vs. Individual) & $-$0.882*** & $-$0.917*** & $-$0.919*** & $-$0.675*** & 0.192*** & 0.210*** & 0.211*** & $-$0.366*** \\
\multicolumn{9}{l}{\textbf{Model Fixed Effects} (vs. GPT-4o mini)} \\
GPT-5.4 & -- & $-$0.080*** & 0.063*** & $-$0.260*** & -- & $-$2.472*** & $-$2.492*** & $-$2.754*** \\
Mistral Large 3 & -- & 1.425*** & 1.517*** & 1.578*** & -- & $-$0.722*** & $-$0.542*** & $-$1.404*** \\
Mistral Medium 3.5 & -- & 0.562*** & 0.734*** & 0.445*** & -- & $-$0.786*** & $-$0.943*** & $-$0.746*** \\
Llama 3.1 8B & -- & $-$0.069*** & $-$0.121*** & 1.645*** & -- & $-$0.846*** & $-$0.566*** & $-$0.895*** \\
Llama 3.1 70B & -- & 0.444*** & 0.655*** & 0.660*** & -- & $-$0.463*** & $-$0.399*** & $-$0.369*** \\
Llama 4 & -- & 0.911*** & 1.018*** & 1.348*** & -- & $-$0.020 & $-$0.043* & $-$0.197*** \\
\multicolumn{9}{l}{\textbf{Model $\times$ Task Structure Interaction}} \\
GPT-5.4 x Asked First & -- & -- & $-$0.455*** & -- & -- & -- & 0.020 & -- \\
Mistral Large 3 x Asked First & -- & -- & $-$0.284*** & -- & -- & -- & $-$0.559*** & -- \\
Mistral Medium 3.5 x Asked First & -- & -- & $-$0.541*** & -- & -- & -- & 0.455*** & -- \\
Llama 3.1 8B x Asked First & -- & -- & 0.173*** & -- & -- & -- & $-$0.871*** & -- \\
Llama 3.1 70B x Asked First & -- & -- & $-$0.657*** & -- & -- & -- & $-$0.201*** & -- \\
Llama 4 x Asked First & -- & -- & $-$0.334*** & -- & -- & -- & 0.068* & -- \\
\multicolumn{9}{l}{\textbf{Model $\times$ Batch/Confidence Interaction}} \\
GPT-5.4 x Batch Prompt & -- & -- & -- & 0.658*** & -- & -- & -- & 0.850*** \\
Mistral Large 3 x Batch Prompt & -- & -- & -- & $-$0.405*** & -- & -- & -- & 1.380*** \\
Mistral Medium 3.5 x Batch Prompt & -- & -- & -- & 0.722*** & -- & -- & -- & 0.117*** \\
Llama 3.1 8B x Batch Prompt & -- & -- & -- & $-$2.308*** & -- & -- & -- & 1.658*** \\
Llama 3.1 70B x Batch Prompt & -- & -- & -- & $-$0.198*** & -- & -- & -- & $-$0.110*** \\
Llama 4 x Batch Prompt & -- & -- & -- & $-$0.379*** & -- & -- & -- & 0.272*** \\
GPT-5.4 x With Confidence & -- & -- & -- & $-$0.348*** & -- & -- & -- & $-$0.430*** \\
Mistral Large 3 x With Confidence & -- & -- & -- & 0.265*** & -- & -- & -- & $-$0.067* \\
Mistral Medium 3.5 x With Confidence & -- & -- & -- & $-$0.478*** & -- & -- & -- & $-$0.243*** \\
Llama 3.1 8B x With Confidence & -- & -- & -- & $-$0.689*** & -- & -- & -- & $-$1.660*** \\
Llama 3.1 70B x With Confidence & -- & -- & -- & $-$0.190*** & -- & -- & -- & $-$0.105*** \\
Llama 4 x With Confidence & -- & -- & -- & $-$0.402*** & -- & -- & -- & 0.081** \\
\multicolumn{9}{l}{\textbf{Random Effects}} \\
$\tau_{00}$\textsubscript{tweet\_id} & 6.771 & 7.222 & 7.253 & 7.695 & 8.916 & 10.092 & 10.172 & 10.547 \\
\midrule
ICC & 0.673 & 0.687 & 0.688 & 0.700 & 0.730 & 0.754 & 0.756 & 0.762 \\
Marginal $R^2$ & 0.024 & 0.049 & 0.051 & 0.069 & 0.002 & 0.045 & 0.048 & 0.058 \\
Conditional $R^2$ & 0.681 & 0.702 & 0.704 & 0.721 & 0.731 & 0.765 & 0.767 & 0.776 \\
AIC & 357253.6 & 344772.5 & 343957.8 & 333283.1 & 480966.9 & 444776.2 & 442366.1 & 432945.0 \\
BIC & 357322.8 & 344911.0 & 344165.4 & 333559.9 & 481036.1 & 444914.6 & 442573.8 & 433221.9 \\
\bottomrule
\end{tabular}%
}
\smallskip
\begin{minipage}{\textwidth}
\raggedright
\footnotesize \textit{Note}: $N$\textsubscript{tweet\_id} = 3,000; observations = 755,999. *** $p<.001$, ** $p<.01$, * $p<.05$.
\end{minipage}
\label{tab:task_structure_glmer}
\end{table*}

\subsection{Model-Specific Task-Design Coefficients}
\label{app:permodel}
Table~\ref{tab:permodel} reports mixed-effects linear probability models fit separately to each LLM. The coefficients show how each task-design choice changes the probability of an OL or HS label within each model. The final column reports the cross-model standard deviation of each estimate.

\begin{table*}[t]
\centering
\caption{\textbf{Model-specific mixed-effects linear probability model estimates for offensive language and hate speech.} Each model column reports a separate model with a random intercept for tweet. The intercept is the labeling probability under separate calls, individual presentation, and no confidence elicitation. The remaining coefficients are changes relative to that condition.}
\resizebox{\textwidth}{!}{%
\begin{tabular}{lccccccc|c}
\toprule
Term & GPT-4o-mini & GPT-5.4 & Mistral Large 3 & Mistral Medium 3.5 & Llama 3.1 8B & Llama 3.1 70B & Llama 4 & SD \\
\midrule
\multicolumn{9}{l}{\textbf{Offensive Language}} \\
Intercept (Separate, individual, no confidence) & .721 & .751 & .831 & .791 & .830 & .802 & .831 & .043 \\
\quad Joint: OL first (vs. Separate) & .068 & .002 & .032 & .003 & .064 & $-$.001 & .020 & .029 \\
\quad Joint: HS first (vs. Separate) & .050 & $-$.022 & .027 & $-$.005 & .016 & .002 & .006 & .023 \\
\quad With Confidence (vs. Without) & .043 & .020 & .046 & .010 & $-$.003 & .028 & .014 & .018 \\
\quad Batch Prompt (vs. Individual) & $-$.045 & $-$.001 & $-$.055 & .003 & $-$.200 & $-$.053 & $-$.059 & .067 \\
ICC (tweet) & .749 & .856 & .697 & .768 & .273 & .727 & .782 & \\
\midrule
\multicolumn{9}{l}{\textbf{Hate Speech}} \\
Intercept (Separate, individual, no confidence) & .434 & .170 & .242 & .340 & .362 & .355 & .349 & .087 \\
\quad Joint: OL first (vs. Separate) & $-$.055 & .014 & .117 & $-$.026 & $-$.014 & .049 & .075 & .061 \\
\quad Joint: HS first (vs. Separate) & $-$.033 & .004 & .004 & .023 & $-$.088 & .001 & .038 & .042 \\
\quad With Confidence (vs. Without) & .020 & $-$.015 & .013 & $-$.002 & $-$.123 & .010 & .027 & .052 \\
\quad Batch Prompt (vs. Individual) & $-$.033 & .036 & .088 & $-$.022 & .110 & $-$.042 & $-$.008 & .061 \\
ICC (tweet) & .848 & .824 & .703 & .742 & .333 & .688 & .791 & \\
\midrule
Observations & 108,000 & 108,000 & 108,000 & 108,000 & 107,999 & 108,000 & 108,000 & \\
\bottomrule
\end{tabular}%
}
\smallskip
\begin{minipage}{\textwidth}
\raggedright
\footnotesize \textit{Note}: $N$\textsubscript{tweet\_id} = 3,000 per model. Coefficients are in probability units; multiply by 10,000 for basis points. Individual coefficients are not starred because the claim concerns the spread across models; pooled tests appear in Table~\ref{tab:taskeffects_all}.
\end{minipage}
\label{tab:permodel}
\end{table*}

\subsection{Agreement Between Task Designs}
\label{app:agreements_appendix_fewshot}
We use Fleiss' $\kappa$ \citep{fleiss1971measuring} to calculate within-design agreement among the three runs and Cohen's $\kappa$ to compare modal labels across designs and against human reference ratings from \citet{kern2023annotations}.

\begin{table*}[t]
\centering
\footnotesize
\setlength{\tabcolsep}{6pt}
\caption{\textbf{Agreement by model.} Min--max (mean) over task designs or design pairs. Within a design is Fleiss' $\kappa$ over three runs; across designs is Cohen's $\kappa$ between modal labels from two designs of the same model; against humans is Cohen's $\kappa$ against the human majority label. The pooled rows give min--max (median).}
\label{tab:agreement_ladder}
\begin{tabular}{@{}lccc@{}}
\toprule
Model & Within a design & Across designs & Against humans \\
 & (Fleiss' $\kappa$) & (Cohen's $\kappa$) & (Cohen's $\kappa$) \\
\midrule
GPT-4o-mini & .84--.97 (.94) & .57--.94 (.79) & .41--.62 (.54) \\
GPT-5.4 & .91--.96 (.93) & .69--.92 (.84) & .50--.66 (.59) \\
Mistral Large 3 & .89--.95 (.92) & .46--.90 (.68) & .22--.60 (.47) \\
Mistral Medium 3.5 & .81--.92 (.87) & .61--.91 (.77) & .36--.67 (.56) \\
Llama 3.1 8B & .19--.80 (.46) & $-$.00--.81 (.35) & .14--.54 (.34) \\
Llama 3.1 70B & .66--.89 (.79) & .62--.89 (.74) & .42--.62 (.54) \\
Llama 4 & .94--.99 (.97) & .56--.92 (.77) & .32--.59 (.50) \\
\midrule
All models & .19--.99 (.91) & $-$.00--.94 (.76) & .14--.67 (.54) \\
Excluding Llama 3.1 8B & .66--.99 (.92) & .46--.94 (.78) & .22--.67 (.56) \\
\bottomrule
\end{tabular}
\end{table*}

\subsection{Confidence Scores by Model}
\label{app:confidence}
\begin{figure*}[t]
    \centering
    \includegraphics[width=\textwidth]{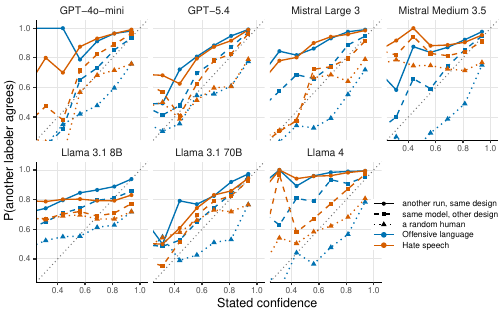}
    \caption{Calibration curves by model. Each panel plots observed agreement against stated confidence for one model, with one line per referent. The diagonal marks perfect calibration; points below it indicate overconfidence.}
    \label{fig:confidence_permodel}
\end{figure*}

Table~\ref{tab:confidence_permodel} reports, for each model and outcome, ECE against three referents, mean stated confidence, the share of items whose modal label changes across designs, and the AUC of confidence as a detector of those changes. Figure~\ref{fig:confidence_permodel} shows the calibration curves.

\begin{table*}[t]
\centering
\footnotesize
\caption{\textbf{Calibration and change detection.} Panel A reports Expected Calibration Error (ECE) pooled across all seven models for the four agreement targets discussed in Section~\ref{sec:confidence}. Panel B reports model-specific ECE for the first three targets, mean stated confidence, the share of items whose modal label changes across designs, and the AUC of confidence as a detector of those changes.}
\label{tab:confidence_permodel}

\textbf{Panel A: Pooled calibration across seven models}

\smallskip
\begin{tabular}{lrrrr}
\toprule
& \multicolumn{4}{c}{ECE against agreement with\ldots} \\
\cmidrule(lr){2-5}
Outcome
& Another run
& Another design
& Individual human
& Human majority \\
\midrule
Offensive language & 0.074 & 0.050 & 0.218 & 0.160 \\
Hate speech         & 0.096 & 0.059 & 0.163 & 0.111 \\
\bottomrule
\end{tabular}

\vspace{0.8em}

\textbf{Panel B: Model-specific calibration and change detection}

\smallskip
\begin{tabular}{lrrrrrr}
\toprule
& \multicolumn{3}{c}{ECE against agreement with\ldots} & & & \\
\cmidrule(lr){2-4}
Model
& Another run
& Another design
& Individual human
& Mean conf.
& Change rate
& AUC \\
\midrule
\multicolumn{7}{l}{\emph{Offensive language}} \\
\quad GPT-4o-mini        & 0.069 & 0.021 & 0.200 & 90.4 & 27.1\% & 0.706 \\
\quad GPT-5.4            & 0.052 & 0.021 & 0.185 & 92.5 & 14.3\% & 0.931 \\
\quad Mistral Large 3    & 0.085 & 0.023 & 0.226 & 89.6 & 21.7\% & 0.907 \\
\quad Mistral Medium 3.5 & 0.027 & 0.030 & 0.230 & 94.0 & 19.8\% & 0.897 \\
\quad Llama 3.1 8B       & 0.173 & 0.172 & 0.247 & 72.6 & 76.1\% & 0.710 \\
\quad Llama 3.1 70B      & 0.060 & 0.052 & 0.230 & 90.4 & 22.6\% & 0.928 \\
\quad Llama 4            & 0.110 & 0.074 & 0.211 & 88.1 & 19.9\% & 0.735 \\
\addlinespace[3pt]
\multicolumn{7}{l}{\emph{Hate speech}} \\
\quad GPT-4o-mini        & 0.111 & 0.059 & 0.130 & 86.3 & 21.8\% & 0.813 \\
\quad GPT-5.4            & 0.034 & 0.007 & 0.175 & 94.4 & 14.6\% & 0.929 \\
\quad Mistral Large 3    & 0.124 & 0.025 & 0.114 & 84.3 & 37.9\% & 0.765 \\
\quad Mistral Medium 3.5 & 0.040 & 0.050 & 0.160 & 89.9 & 29.5\% & 0.741 \\
\quad Llama 3.1 8B       & 0.232 & 0.225 & 0.250 & 66.4 & 72.1\% & 0.683 \\
\quad Llama 3.1 70B      & 0.094 & 0.089 & 0.194 & 81.7 & 35.5\% & 0.820 \\
\quad Llama 4            & 0.162 & 0.065 & 0.141 & 82.0 & 31.0\% & 0.869 \\
\bottomrule
\end{tabular}

\smallskip
\begin{minipage}{\textwidth}
\raggedright
\footnotesize
\textit{Note}: ECE uses ten equal-width bins of stated confidence. ``Another run'' is a second run of the same model and task design. ``Another design'' is the same model applied under a different task design. ``Individual human'' is a randomly drawn human annotator, and ``human majority'' is the majority human label. Mean confidence is reported on the 0--100 scale. Change rate is the share of items whose modal label changes across task designs. AUC measures how well lower mean item confidence identifies those changes.
\end{minipage}
\end{table*}

\subsection{Annotation Costs}
\label{app:costs}
Batch presentation reduces API calls sixfold and cost by approximately 76\% within each model. GPT-5.4 costs approximately 17 times more than GPT-4o-mini for the same 3,000 tweets.

\begin{table}[h]
\centering
\footnotesize
\caption{\textbf{Annotation costs for three runs of 3,000 tweets.} Batch presentation reduces API calls sixfold and cost by approximately 76\% within each model; GPT-5.4 costs approximately 17 times more than GPT-4o-mini.}
\label{tab:costs}
\setlength{\tabcolsep}{2pt}
\begin{tabular}{lcccccc}
\toprule
& \textbf{API} & \multicolumn{3}{c}{\textbf{Tokens per call}} & \multicolumn{2}{c}{\textbf{Cost (\$), 3 runs}} \\
\cmidrule(lr){3-5}\cmidrule(lr){6-7}
\textbf{Condition} & \textbf{Calls/run} & \textbf{Task} & \textbf{Tweet} & \textbf{Out} & \textbf{4o-mini} & \textbf{5.4} \\
\midrule
Indiv., no conf. & 3000 & 517 & 19 & 2 & 0.73 & 12.33 \\
Indiv., w/ conf. & 3000 & 580 & 19 & 4 & 0.83 & 14.02 \\
Batch, no conf. & 500 & 639 & 112 & 12 & 0.18 & 3.09 \\
Batch, w/ conf. & 500 & 724 & 112 & 24 & 0.21 & 3.67 \\
\bottomrule
\end{tabular}
\end{table}

\end{document}